\pdfoutput=1

\documentclass[11pt]{article}
\usepackage{enumitem}   

\usepackage[final]{acl}
\usepackage{multirow}
\usepackage{booktabs}
\usepackage{times}
\usepackage{latexsym}
\usepackage{amsmath} 
\usepackage{amsthm}           
\theoremstyle{definition}     
\newtheorem{definition}{Definition}[section]  
\usepackage[T1]{fontenc}

\usepackage[utf8]{inputenc}

\usepackage{microtype}

\usepackage{inconsolata}

\usepackage{graphicx}
\usepackage[ruled,vlined,linesnumbered]{algorithm2e}
\usepackage{amssymb}
\usepackage[most]{tcolorbox}
\tcbuselibrary{listings, breakable}
\usepackage{enumitem}

\newcounter{prompt}

\title{SchemaGraphSQL: Efficient Schema Linking with Pathfinding Graph Algorithms for Text-to-SQL on Large-Scale Databases}

\author{
\textbf{AmirHossein Safdarian}$^{1}$, \textbf{Milad Mohammadi}$^{1}$, \textbf{Ehsan Jahanbakhsh}$^{1}$, \\
\textbf{Mona Shahamat Naderi}$^{2}$, \textbf{Heshaam Faili}$^{1}$ \\
$^{1}$University of Tehran, Iran \quad
$^{2}$Sharif University of Technology, Iran \\
\texttt{\{a.safdarian, miladmohammadi, ehsan.jahanbakhsh, hfaili\}@ut.ac.ir}, \\
\texttt{mona.shahamat@sharif.edu}
}

\begin{document}
\maketitle
\begin{abstract}
Text-to-SQL systems translate natural language questions into executable SQL queries, and recent progress with large language models (LLMs) has driven substantial improvements in this task. Schema linking remains a critical component in Text-to-SQL systems, reducing prompt size for models with narrow context windows and sharpening model focus even when the entire schema fits. We present a zero-shot, training-free schema linking approach that first constructs a schema graph based on foreign key relations, then uses a single prompt to Gemini 2.5 Flash to extract source and destination tables from the user query, followed by applying classical path-finding algorithms and post-processing to identify the optimal sequence of tables and columns that should be joined, enabling the LLM to generate more accurate SQL queries. Despite being simple, cost-effective, and highly scalable, our method achieves state-of-the-art results on the BIRD benchmark, outperforming previous specialized, fine-tuned, and complex multi-step LLM-based approaches. We conduct detailed ablation studies to examine the precision–recall trade-off in our framework. Additionally, we evaluate the execution accuracy of our schema filtering method compared to other approaches across various model sizes.
\end{abstract}

\section{Introduction}

Relational databases are foundational to modern data infrastructure, powering analytics, reporting, and decision-making across domains. Yet, querying these databases typically requires fluency in SQL—a barrier for many users. Text-to-SQL systems aim to democratize access by translating natural language (NL) questions into executable SQL queries~\cite{zhu2024largelanguagemodelenhanced,zhang2024benchmarkingtexttosqlcapabilitylarge}. Enabled by large language models (LLMs), recent systems achieve impressive performance across complex cross-domain settings.

\begin{algorithm}[h]
\footnotesize
\caption{\fontsize{10pt}{10pt}\selectfont Graph-Based Schema Linking}
\KwIn{Question $q$; schema graph $G$;}
\KwOut{Relevant table set $\mathcal{T}^{\star}\subseteq\mathcal{T}$}
\BlankLine
\BlankLine
\textbf{Step 1: Identify source/destination tables}\;
\Indp
$(\mathcal{T}_{src},\mathcal{T}_{dst}) \leftarrow \operatorname{LLM\_call}(q)$
\BlankLine
\Indm

\textbf{Step 2: Build candidate path set}\;
\Indp
$\mathcal{C} \leftarrow \varnothing$\;
\ForEach{$T_{src} \in \mathcal{T}_{src}$, $T_{dst} \in \mathcal{T}_{dst}$}{
    $\mathcal{C} \leftarrow \mathcal{C} \cup \operatorname{ShortestPaths}(T_{src},T_{dst})$
}
\Indm

\BlankLine
\textbf{Step 3: Build union path}\;
\Indp
$U \leftarrow \bigcup_{p \in \mathcal{C}} p$\;
\Return $U$\;
\Indm
\end{algorithm}
However, bringing these systems to real-world applications introduces new challenges. Enterprise databases often contain hundreds of tables and thousands of columns—far beyond the scale of academic benchmarks. Supplying the entire schema to the model risks exceeding token limits and introduces considerable noise, which can hinder SQL generation and inflate inference cost~\cite{cao2024rslsqlrobustschemalinking,li2023llmservedatabaseinterface}. In practice, user queries typically touch only a small subset of the schema, making it crucial to identify and extract the relevant part—a process known as \emph{schema linking}~\cite{lei-etal-2020-examining}.

Schema linking aims to determine which tables or columns are needed to answer a user question. While early methods relied on exact string matches~\cite{yu-etal-2018-typesql}, recent work has proposed neural linkers~\cite{gan-etal-2023-appraising}, retrieval-based modules~\cite{pourreza2024dtssqldecomposedtexttosqlsmall}, and prompt-based systems~\cite{wang2025linkalignscalableschemalinking}. These can capture semantic signals beyond surface overlap, but typically require supervised training, complex multi-stage pipelines, or brittle prompt engineering. They also struggle with the core trade-off: being precise enough to reduce noise, yet broad enough not to miss critical context~\cite{liu2024solidsqlenhancedschemalinkingbased,wang2025dbcopilotnaturallanguagequerying}.

\vspace{0.5em}
\textbf{In this work, we ask: \emph{Can we perform effective schema linking without relying on specialized fine-tuned models or complex prompting strategies?}} Our answer is affirmative.

We introduce \textbf{SchemaGraphSQL}, a zero-shot schema linking framework that revisits classical algorithmic tools. Our key idea is to model schema linking as a graph search problem. We treat the database schema as a graph where nodes are tables and edges reflect foreign-key connections. Given a user query, we make a single LLM call to predict coarse-grained source and destination tables, then apply deterministic path-finding algorithms to enumerate all shortest join paths between them. The union of these paths forms a compact sub-schema—guaranteed to be connected and grounded in the query.

This perspective is both simple and surprisingly powerful. \textbf{To our knowledge, SchemaGraphSQL is the first Text-to-SQL system to rely exclusively on classical graph algorithms for schema linking, using LLMs only for coarse guidance.} It requires no training, incurs minimal inference cost, and integrates easily into any downstream parser or LLM-based SQL generator.

Empirical results on the BIRD benchmark show that SchemaGraphSQL achieves new state-of-the-art scores on recall-focused schema linking metrics and improves execution accuracy across multiple SQL generators. We also conduct ablations demonstrating that even this minimal linking method outperforms specialized neural or prompt-based systems in robustness and cost-efficiency.

\begin{figure*}[t]
  \centering
  \includegraphics[width=0.9\linewidth]{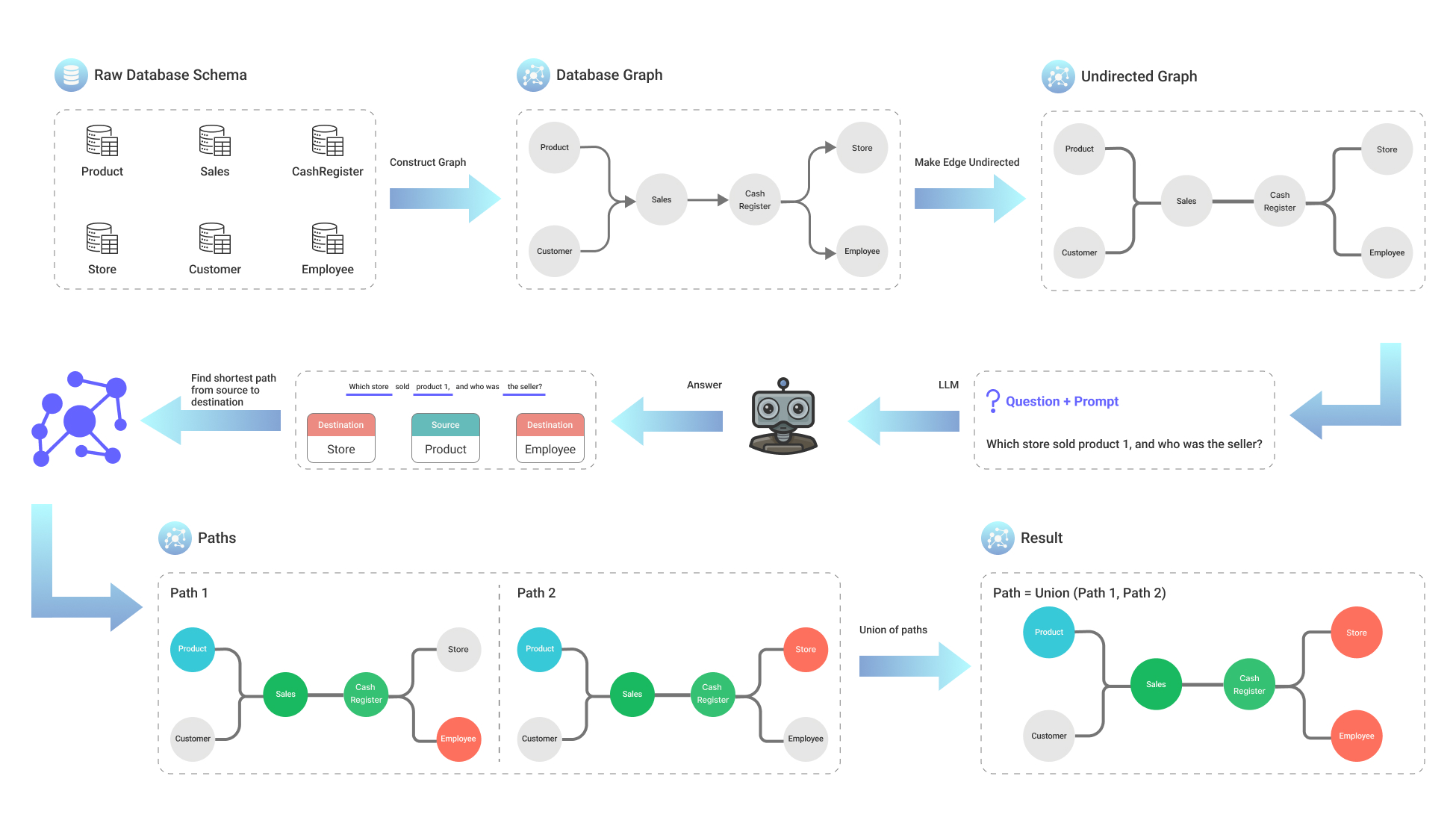}
  \caption{Overview of our graph-based schema linking pipeline.}
  \label{fig:graph-method}
\end{figure*}

\textbf{Main Contributions:}
\begin{itemize}
\item We introduce a zero-shot schema linking approach that models database schemas as graphs and applies classical path-finding algorithms. Our method achieves state-of-the-art performance without requiring any training—either for fine-tuning or inference—making it highly suitable for low-resource, real-world scenarios where training data is unavailable or difficult to obtain.
\item Our system uses only a single lightweight LLM call (Gemini 2.5 Flash) per query, with minimal token usage (averaging 4593 input and 14 output tokens), significantly reducing inference cost while maintaining ease of integration and deployment.
\item We conduct comprehensive empirical evaluations, demonstrating superior schema linking performance compared to fine-tuned and specialized methods. Additionally, we perform detailed ablation studies to examine precision–recall trade-offs and assess the downstream impact on Text-to-SQL execution accuracy across a range of open-source and closed-source models.
\end{itemize}

\section{Related Work}
\label{sec:related}

Text-to-SQL systems aim to automatically translate natural language questions into executable SQL queries, thereby enabling non-experts to interact with relational databases. The advent of large language models (LLMs) has significantly advanced this task~\cite{zhang2024benchmarkingtexttosqlcapabilitylarge,zhu2024largelanguagemodelenhanced}, with models like GPT-3.5/4, Gemini, and their open-source variants demonstrating impressive performance across benchmarks. However, as schema size increases, providing the entire schema as input may exceed the model’s context window, especially in large-scale databases. Even when using recent LLMs with extended context lengths, supplying the full schema can introduce noise and hinder the model’s ability to focus on relevant elements.

\subsection{Schema Linking in Text-to-SQL} 
Schema linking—the process of aligning natural language mentions to corresponding tables and columns in a database—is a crucial component of Text-to-SQL systems~\cite{lei-etal-2020-examining,10.1145/3534678.3539294,li2023llmservedatabaseinterface}. Early approaches relied on exact string matching or type-based heuristics~\cite{yu-etal-2018-typesql}, which struggled with synonyms, paraphrases, and complex cross-domain schemas. Recent methods have increasingly leveraged pretrained LLMs and neural encoders to improve linking accuracy~\cite{gan-etal-2023-appraising,glass2025extractiveschemalinkingtexttosql}. Schema linking has proven particularly important for LLM pipelines that operate on large or multi-database environments, where prompt space is limited and precision in schema filtering directly affects SQL generation quality~\cite{cao2024rslsqlrobustschemalinking,liu-etal-2025-solid}.

\subsection{Neural and Prompt-Based Linking Strategies} 
Numerous methods have been proposed to handle schema linking within LLM-based Text-to-SQL systems. Some decouple schema linking as a separate module before SQL generation~\cite{pourreza2024dtssqldecomposedtexttosqlsmall,Li2023RESDSQLDS}, while others incorporate schema selection as a prompt-driven or retrieval-augmented step~\cite{wang2025linkalignscalableschemalinking}. Extractive methods, such as~\citet{glass2025extractiveschemalinkingtexttosql}, directly prompt LLMs to list relevant schema items, trading generation flexibility for interpretability and control. RSL-SQL~\cite{cao2024rslsqlrobustschemalinking} proposes a bidirectional pruning mechanism with self-correction to boost recall, while Solid-SQL~\cite{liu-etal-2025-solid} augments training data to improve linking robustness. Despite variations in architecture, a common trend across these systems is the effort to balance schema coverage (recall) with relevance filtering (precision) to avoid overloading the LLM or omitting critical elements.

\subsection{Graph-Based Approaches for Schema Linking} 
A parallel line of work models the database schema as a graph structure, where tables and columns are nodes, and foreign-key or semantic relations form edges. These methods primarily leverage graph neural networks (GNNs) or relation-aware transformers to propagate information across schema components. RAT-SQL~\cite{wang-etal-2020-rat} pioneered relation-aware attention over a joint question–schema graph, inspiring successors such as LGESQL~\cite{cao-etal-2021-lgesql} (line-graph encoding of meta-relations) and ShadowGNN~\cite{chen-etal-2021-shadowgnn} (delexicalised projection for cross-schema generalisation).  
Later hybrids integrate graph reasoning directly into pretrained LMs, e.g.\ Graphix-T5~\cite{li2023graphixt5mixingpretrainedtransformers} and GRL-SQL~\cite{10.1145/3626772.3657961}.  
Most recently, SQLformer~\cite{bazaga2024sqlformerdeepautoregressivequery} embeds schema structure as inductive bias in a Transformer encoder and autoregressively generates SQL ASTs as graphs.  
While graph‐enhanced models capture rich global relations, they typically require substantial fine-tuning or architectural changes—an obstacle in low-resource, real-time deployments. Graph-based schema linking methods have recently declined in popularity as LLM-driven approaches have become dominant.

\subsection{Classical Graph Algorithms in Schema Linking} 
In contrast to learned graph encoders, only a handful of systems reuse \emph{classical} graph algorithms to aid LLMs.  
\textbf{DBCopilot}~\cite{wang2025dbcopilotnaturallanguagequerying} constructs a directed schema graph and performs depth-first traversal to linearise the sub-schema passed to a lightweight “router” model.  
\textbf{Interactive-T2S}~\cite{xiong2024interactivet2smultiturninteractionstexttosql} equips an LLM agent with a \textsc{FindShortestPath} tool that runs breadth-first search over the foreign-key graph to supply valid join chains during multi-turn dialogue.  
These works demonstrate the practicality of DFS/BFS as auxiliary helpers, but the graph search remains peripheral—responsible only for join validation or routing—rather than serving as the \emph{core} schema-linking engine.

\subsection{Positioning Our Work} 
While prior literature has thoroughly explored neural and graph-enhanced architectures for schema linking, the explicit use of classical graph algorithms—particularly as the \textit{core mechanism} for schema linking in LLM-based Text-to-SQL systems—remains rare. Our approach, \textsc{SchemaGraphSQL}, revisits this paradigm by operationalizing schema linking as a path-selection problem on the schema graph. To our knowledge, this is the first work to systematically evaluate and ablate classic path-finding algorithms for schema linking in LLM-driven Text-to-SQL pipelines on real-world benchmarks.

\section{Methodology}
\label{sec:method}
\begin{tcolorbox}[colback=gray!6!white, colframe=gray!80!black, title={\textbf{Notation}}]
\textbf{Databases.}
A \emph{relational database} is represented as
\[
\mathcal{D} = \langle \mathcal{T}, \mathcal{A}, \mathcal{K} \rangle,
\]
where:
\vspace{-0.5em}
\begin{itemize}[leftmargin=1.6em]
  \item $\mathcal{T} = \{T_1, \dots, T_n\}$: set of tables.
  \item $A(T_i)$: attributes (columns) of table $T_i$; $\mathcal{A} = \bigcup_{T_i \in \mathcal{T}} A(T_i)$ is the global set of attributes.
  \item $\mathcal{K} \subseteq \mathcal{T} \times \mathcal{T}$: set of foreign key (FK) relations.
\end{itemize}
The \emph{schema graph} is the undirected graph $G = (\mathcal{T}, \mathcal{K})$, with nodes as tables and edges as FK links. For sparse schemas (fewer than two edges), we further augment the schema graph by adding edges between tables that share a column containing ``id'' in its name, thus ensuring that the schema graph is sufficiently connected for path enumeration.

\medskip

\textbf{Languages.}
\begin{itemize}[leftmargin=1.6em]
  \item $\mathcal{L}$: set of well-formed natural language questions.
  \item $\mathcal{S}$: set of valid SQL queries.
\end{itemize}
Given $q \in \mathcal{L}$, the objective is to generate $Q \in \mathcal{S}$ that answers $q$ over $\mathcal{D}$.
\end{tcolorbox}
This section formalizes the schema linking problem and describes our graph-based, training-free approach for selecting minimal connected sub-schemas to facilitate Text-to-SQL generation. We begin by introducing notation and the problem formulation, then present our graph-based schema linking procedure, and finally detail the configuration space of our approach.

\subsection{Problem Formulation}

We first introduce the notation used throughout this paper:

\begin{definition}[Text-to-SQL]
\label{def:t2sql}
Given $q$ and $\mathcal{D}$, \textbf{Text-to-SQL} seeks a function
\begin{center}
\fbox{%
$\displaystyle
f_{\mathrm{NL2SQL}}\;:\;\mathcal{L}\times\mathcal{D}\longrightarrow\mathcal{S}
$%
}
\end{center}
that returns an executable SQL query $Q = f_{\mathrm{NL2SQL}}(q,\mathcal{D})$ that answers the user question $q$ on the database $\mathcal{D}$.
\end{definition}

\begin{definition}[Schema Linking]
\label{def:schemalink}
Let $G=(\mathcal{T},\mathcal{K})$ be the schema graph of $\mathcal{D}$.
\textbf{Schema linking} selects a connected sub-schema $S=\langle\mathcal{T}^\star,\mathcal{K}^\star\rangle$ with $\mathcal{T}^\star\subseteq\mathcal{T}$ and $\mathcal{K}^\star\subseteq\mathcal{K}$ sufficient to express the SQL query answering $q$. Formally,
\begin{center}
\fbox{%
$\displaystyle
g_{\mathrm{SL}}\;:\;\mathcal{L}\times G \longrightarrow
\mathcal{P}(\mathcal{T}),\qquad
\mathcal{T}^\star = g_{\mathrm{SL}}(q,G)
$%
}
\end{center}
Here,
\begin{center}
\fbox{%
$\displaystyle
\mathcal{K}^\star = \{(T_i, T_j) \in \mathcal{K} \mid T_i, T_j \in \mathcal{T}^\star\}
$%
}
\end{center}
The output sub-schema $S$ defines the smallest set of tables and links needed to answer $q$ while remaining connected within the schema graph.
\end{definition}

\subsection{Graph-Based Schema Linking as Path Selection}
\label{ssec:graph_method}

\paragraph{Step 1: Extracting Source and Destination Tables.}
A single LLM call extracts two subsets of tables from the schema:
\begin{itemize}[leftmargin=1.5em]
  \item $\mathcal{T}_s$ (\emph{sources}): tables whose columns appear in query conditions or filtering predicates;
  \item $\mathcal{T}_d$ (\emph{destinations}): tables containing the columns requested as output.
\end{itemize}
Both sets are guaranteed to be non-empty and may overlap, reflecting cases where the same table is used for both filtering and output.

We operationalize schema linking as a path-selection task on the schema graph $G$, which enables systematic and efficient sub-schema identification:

This extraction is performed via a single call to Gemini 2.5 Flash, guided by a dedicated system prompt designed to elicit precise identification of source and destination tables from the question and schema. The full prompt is shown in Prompt~\ref{prompt:srcdst}.

\refstepcounter{prompt}
\begin{tcolorbox}[
    enhanced,
    breakable,
    width=\linewidth,
    colback=gray!5,
    colframe=blue!60!black,
    fonttitle=\bfseries,
    title={\textbf{Prompt~\arabic{prompt}: System prompt for source and destination extraction}},
    boxrule=0.6pt,
    arc=2pt,
    left=2pt, right=2pt, top=2pt, bottom=2pt
  ]
\footnotesize
\textbf{ROLE \& OBJECTIVE}\\
You are a senior data engineer who analyses SQL schemas and maps user questions precisely to
\emph{source} tables (filtering) and \emph{destination} tables (final result columns).

\medskip
\textbf{TASK}\\
Identify:
\begin{itemize}[leftmargin=1.1em,itemsep=2pt]
  \item \textbf{Source table(s) \texttt{(src):}} contain columns used in filters/conditions.
  \item \textbf{Destination table(s) \texttt{(dst):}} contain columns returned in the answer.
\end{itemize}

\medskip
\textbf{INSTRUCTIONS}
\begin{enumerate}[leftmargin=1.2em,itemsep=3pt]
  \item Internally inspect every table to determine  
        \begin{itemize}[leftmargin=*,nosep]
        \item which tables participate in filtering, and  
        \item which tables supply the requested output columns.
        \end{itemize}
        Briefly justify your choice \emph{internally} but \emph{do not} include that justification in the final answer.
  \item Output exactly one line in the following format:  
        \texttt{src=TableA,TableB}, \texttt{dst=TableC,TableD}
\end{enumerate}
\end{tcolorbox}
\label{prompt:srcdst}
\paragraph{Step 2: Candidate Path Enumeration.}
For every pair $(T_s, T_d) \in \mathcal{T}_s \times \mathcal{T}_d$, we enumerate all shortest simple paths connecting them in $G$:
\begin{center}
\fbox{%
\parbox{0.98\linewidth}{%
$\displaystyle
\operatorname{SP}(T_s, T_d) = \Big\{\, p\ \Big|\ p\ \text{is a simple path } T_s \leadsto T_d,\,\, |p| = \operatorname{dist}_G(T_s, T_d) \Big\}
$%
}
}
\end{center}
This set $\operatorname{SP}(T_s, T_d)$ contains all minimal-length paths in the schema graph between each source and destination table pair.

The global candidate set and their union are defined as:
\begin{center}
\fbox{%
$\displaystyle
\mathcal{C} = \bigcup_{T_s\in\mathcal{T}_s}\,
              \bigcup_{T_d\in\mathcal{T}_d}\operatorname{SP}(T_s,T_d), \qquad
U = \bigcup_{p\in\mathcal{C}} p
$%
}
\end{center}
Here, $\mathcal{C}$ enumerates all candidate paths, and $U$ is the union of all tables appearing in any candidate path—representing the maximal connected subgraph that could be relevant for the query.

\paragraph{Step 3: Path Selection and Sub-schema Construction.}
Depending on the configuration (detailed below), the set $U$ is optionally appended to $\mathcal{C}$. A second LLM call (or a deterministic rule) selects a candidate path $p^\star \in \mathcal{C}$, and we set $\mathcal{T}^\star := p^\star$ as the chosen subset of relevant tables for downstream SQL generation.

\subsection{Configurations}
\label{ssec:configs}

To provide flexibility and support empirical analysis, we define a family of selection strategies parameterized by the following flags: let $k_s = |\mathcal{T}_s| > 0$, $k_d = |\mathcal{T}_d| > 0$,
\[
\begin{aligned}
\textsc{Longest} &\in \{\mathsf{false}, \mathsf{true}\}, \\
\textsc{Union} &\in \{\mathsf{false}, \mathsf{true}\}.
\end{aligned}
\]

Table~\ref{tab:configs} summarizes the seven configurations we evaluate, spanning single-source/single-destination and union-based settings.

\begin{center}
\begin{tabular}{@{}lccc@{}}
\toprule
\# & $(k_s,k_d)$ & \textsc{Longest} & \textsc{Union} \\
\midrule
1 & $(1,1)$ & \textsf{false} & \textsf{true} \\
2 & $(1,\ast)$ & \textsf{false} & \textsf{true} \\
3 & $(\ast,1)$ & \textsf{false} & \textsf{true} \\
4 & $(\ast,\ast)$ & \textsf{false} & \textsf{true} \\
5 & $(\ast,\ast)$ & \textsf{true}  & \textsf{true} \\
6 & $(\ast,\ast)$ & \textsf{false} & \textsf{false} \\
7 & $(\ast,\ast)$ & \textsf{false} & always select $U$ \\
\bottomrule
\label{tab:configs}
\end{tabular}
\end{center}

\noindent
Here, $\ast$ means any positive integer. Mode 5 chooses the longest among the shortest paths; Mode 6 excludes $U$ from $\mathcal{C}$; Mode 7 bypasses path selection and deterministically returns the union $U$. This design enables ablation studies to assess the effect of schema coverage and path selection criteria on final Text-to-SQL accuracy.

\subsection{End-to-End Objective}

Given configuration $\Theta$, our full pipeline is:
\begin{center}
\fbox{%
$\displaystyle
f_{\mathrm{NL2SQL}}^{\Theta}(q,\mathcal{D})
= h_{\mathrm{GEN}}\bigl(q,\ g_{\mathrm{SL}}^{\Theta}(q,G)\bigr)
$%
}
\end{center}
where $g_{\mathrm{SL}}^{\Theta}$ is our graph-based schema linker and $h_{\mathrm{GEN}}$ is any downstream SQL generator, constrained to use only the filtered schema $\mathcal{T}^\star$. All pipeline steps operate in a single pass, are fully automatic, and require no training data or domain adaptation.

\section{Experimental Setup}
\label{sec:exp}

\subsection{Dataset}
\label{ssec:dataset}

All experiments are conducted on the \textsc{BIRD} development split, which comprises 1,534 natural-language questions over 11 heterogeneous relational databases.
For schema linking precision, recall, and exact match rate, we use the BIRD dev set gold queries by extracting the referenced tables.
For execution accuracy, we follow the official evaluation script provided by \textsc{BIRD} without modification.

\subsection{Compared Methods}
\label{ssec:baselines}

\paragraph{SchemaGraphSQL (Ours)}
Unless otherwise noted, results correspond to \emph{Mode~7} in Table~\ref{tab:configs}, i.e., we deterministically return the union~$U$ of all shortest paths connecting the LLM-identified source and destination tables (cf.\ Section~\ref{ssec:graph_method}).  
The \texttt{src}/\texttt{dst} extraction prompt (Prompt~\ref{prompt:srcdst}) is executed using \texttt{google/gemini-2.5-flash-preview} at temperature~0.2, while downstream SQL generation is performed at temperature~0.3.

\paragraph{LLM as Schema Linker (Baseline)}
A single Gemini~2.5 Flash call is prompted to list \emph{all} tables that must appear in the \texttt{FROM}/\texttt{JOIN} clause given the user question.  
This mirrors prior “single-step” schema linking approaches while controlling for model and prompt length.

\paragraph{\textsc{Dense Retriever}}
We embed each table name (along with its column names) using the multilingual-\textsc{E5}-large-instruct encoder.  
For each question, the top-$k$ tables ($k = 1\ldots6$) retrieved via cosine similarity form the predicted schema.

\paragraph{}
For completeness, we also include published \textsc{BIRD} dev results from recent schema-linking systems such as Extractive Schema Linking for Text-to-SQL \cite{glass2025extractiveschemalinkingtexttosql} and \textsc{LinkAlign}.  
\cite{wang2025linkalignscalableschemalinking} We did \emph{not} re-run these systems; hence, they are excluded from execution accuracy comparisons.

\subsection{LLMs for SQL Generation}
\label{ssec:llms}

Following schema filtering, we evaluate four LLMs for SQL generation:

\begin{itemize}
  \item \texttt{google/gemini-2.5-flash-preview};
  \item \texttt{google/gemma-3-27b-it};
  \item \texttt{google/gemma-3-12b-it};
  \item \texttt{google/gemma-3-4b-it}.
\end{itemize}

All calls are made through the respective provider APIs using identical configurations and prompting templates.

\subsection{Evaluation Metrics}
\label{ssec:metrics}

\paragraph{Schema-level Metrics.}
Let $G$ be the gold table set and $P$ the predicted set.

\begin{itemize}[leftmargin=1.5em]
  \item \textbf{Precision:} The percentage of predicted tables that are actually present in the gold SQL query:
  \[
  \text{Precision} = \frac{|P \cap G|}{|P|}
  \]
  
  \item \textbf{Recall:} The percentage of gold tables that are successfully predicted:
  \[
  \text{Recall} = \frac{|P \cap G|}{|G|}
  \]

  \item \textbf{$F_\beta$ Score:} The generalized F-score that weights recall $\beta$ times more than precision:
  \[
  F_{\beta} = \frac{(1+\beta^2)\,|P \cap G|}{\beta^2|G| + |P|}, 
  \qquad
  \beta \in \{1, 6\}
  \]

  \item \textbf{Exact Match Rate (EMR):} The percentage of examples where the predicted schema exactly matches the gold schema:
  \[
  \text{EMR} = \frac{1}{N} \sum_{i=1}^{N} \mathbb{I}[P_i = G_i]
  \]
  
\end{itemize}
\paragraph{End-to-End Metric}
\emph{Execution accuracy} is computed using the official \textsc{BIRD} evaluation script: the generated SQL query is executed against the database, and its result must exactly match that of the reference query.

\subsection{Implementation Notes}
\label{ssec:impl}

All experiments are conducted via hosted API endpoints; no on-premise hardware is used.  
Each query incurs (i) one Gemini~2.5 Flash call for schema linking, and (ii) one model call for SQL generation (Gemini2.5 or Gemma3).  
Code, prompts, and outputs will be released to support reproducibility.

\section{Results}
\label{sec:results}
\begin{table*}[h!]
    \centering
    \caption{Schema Linking Results in Dev Mode}
    \label{tab:schema_linking_dev}
    \resizebox{\linewidth}{!}{
        \begin{tabular}{|l|ccccc|}
            \hline
            \textbf{Method} & \textbf{Exact Match Rate (\%)} & \textbf{Precision (\%)} & \textbf{Recall (\%)} & \textbf{F1 (\%)} & \textbf{F6 (\%)} \\ \hline
            LLM as Schema Linker & 75.88 & 91.79 & 89.90 & 90.83 & 89.95 \\ 
            Retrieval (Top1) & 20.08 & 86.70 & 44.46 & 58.78 & 45.05 \\ 
            Retrieval (Top2) & 26.79 & 66.59 & 67.80 & 67.19 & 67.77 \\ 
            Retrieval (Top3) & 4.63 & 53.67 & 80.91 & 64.54 & 79.82 \\ 
            Retrieval (Top4) & 1.24 & 45.79 & 87.64 & 60.15 & 85.52 \\ 
            Retrieval (Top5) & 1.04 & 39.89 & 91.11 & 55.49 & 88.06 \\ 
            Retrieval (Top6) & 1.04 & 35.43 & 93.31 & 51.36 & 89.37 \\ 
            DIN-SQL & - & 79.90 & 55.70 & 65.64 & 56.16 \\ 
            PET-SQL & - & 81.60 & 64.90 & 72.30 & 65.26 \\ 
            MAC-SQL & - & 76.30 & 56.20 & 64.73 & 56.60 \\ 
            MCS-SQL & - & 79.60 & 76.90 & 78.23 & 76.97 \\ 
            RSL-SQL & - & 78.10 & 77.50 & 77.80 & 77.52 \\ 
            LinkAlign Agent & - & 77.10 & 79.40 & 78.23 & 79.34 \\ 
            DTS-SQL & - & 95.07 & 92.74 & 93.89 & 92.80 \\ 
            Gen & - & 90.40 & 95.50 & 92.88 & 95.35 \\ 
            ExSL$_c$ & - & 95.86 & 93.94 & 94.89 & 93.99 \\ 
            ExSL$_f$ & - & \textbf{96.35} & 93.85 & \textbf{95.08} & 93.92 \\ \hline
            SchemaGraphSQL$_{1-1}$ & 71.06 & 94.89 & 84.02 & 89.12 & 84.28 \\ 
            SchemaGraphSQL$_{force-union}$ & \textbf{76.60} & 86.21 & \textbf{95.71} & 90.71 & \textbf{95.43} \\ 

\hline
        \end{tabular}
    }
\end{table*}

\subsection{Schema Linking Evaluation}
\label{ssec:schema}

Table~\ref{tab:schema_linking_dev} shows that our primary configuration, \textbf{SchemaGraphSQL\textsubscript{force-union}}, attains \textbf{Recall = 95.71\,\%} and an \textbf{F\textsubscript{6}=95.43\,\%} on the \textsc{BIRD} development split—surpassing all published systems, including the previous recall-centric leader ExSL\textsubscript{f} (\textbf{F\textsubscript{6}=93.92\,\%}). Prior work has argued that recall-weighted metrics such as F\textsubscript{6} are the most reliable indicator of downstream success, because omitting a relevant table is far more damaging than including extras~\cite{glass2025extractiveschemalinkingtexttosql}.  By pushing both recall and F\textsubscript{6} to new highs without any supervised training, \textbf{SchemaGraphSQL\textsubscript{force-union}} establishes a new performance bar for zero-shot schema linking.
\paragraph{}
For users who require a tighter schema, our balanced \textbf{SchemaGraphSQL\textsubscript{$n$-$n$}} variant delivers the best F\textsubscript{1} (92.93\,\%) with only a modest drop in recall (95.10\,\%).  Exact-match rate also improves over the single-step LLM baseline (75.88 \%)—rising to 78.29 \% for $n$-$n$ and 76.60 \% for force-union—demonstrating that classical graph search repairs connectivity errors that an LLM alone often misses.

\subsection{Ablation Insights}
\label{ssec:ablation}
\setlength{\tabcolsep}{3pt}
The configuration sweep in Table~\ref{tab:schema_linking_results_methods_dev} highlights two actionable lessons:

\begin{itemize}[leftmargin=1.3em]
    \item \textbf{Union is essential.} Removing the union step (\textit{no-union}) drops both F\textsubscript{1} and EMR, confirming that coverage matters more than compactness.
    \item \textbf{Avoid unnecessary hops.} Forcing the longest path (\textit{force-longest}) harms all metrics, indicating that extra intermediate tables add noise without benefit.
\end{itemize}

Together, these results validate our design choice: merge all shortest paths for maximum recall, then optionally down-select (e.g., $n$-$n$) when higher precision is required.
\begin{table}[ht]
    \centering
    \caption{Schema-linking results across graph settings on \textsc{BIRD}-Dev.}
    \label{tab:schema_linking_results_methods_dev}
    \footnotesize            
    \begin{tabular*}{\columnwidth}{@{\extracolsep{\fill}} lccccc}
        \toprule
        \textbf{Method} & \textbf{EMR} & \textbf{Prec.} & \textbf{Rec.} & \textbf{F1} & \textbf{F6} \\
                        & (\%)        & (\%)          & (\%)          & (\%)        & (\%)        \\
        \midrule
        \tiny{SchemaGraphSQL$_{1-1}$}                        & 71.06 & \textbf{94.89} & 84.02 & 89.12 & 84.28 \\
        \tiny{SchemaGraphSQL$_{1-n}$}                        & 78.16 & 93.29 & 91.55 & 92.41 & 91.60 \\
        \tiny{SchemaGraphSQL$_{n-1}$}                        & 78.23 & 90.99 & 94.86 & 92.89 & 94.76 \\
        \tiny{SchemaGraphSQL$_{n-n}$}                        & \textbf{78.29} & 90.87 & 95.10 & \textbf{92.93} & 94.98 \\
        \tiny{SchemaGraphSQL$_{\mathit{force\text{-}longest}}$} & 71.64 & 89.47 & 88.45 & 88.96 & 88.47 \\
        \tiny{SchemaGraphSQL$_{\mathit{no\text{-}union}}$}      & 73.73 & 91.39 & 90.03 & 90.71 & 90.07 \\
        \tiny{SchemaGraphSQL$_{\mathit{force\text{-}union}}$}   & 76.60 & 86.21 & \textbf{95.71} & 90.71 & \textbf{95.43} \\
        \bottomrule
    \end{tabular*}
\end{table}
\begin{table*}[h!]
    \centering
    \caption{SQL Execution Accuracy Results - Dev}
    \label{tab:execution_accuracy_dev}
    \resizebox{\linewidth}{!}{
        \begin{tabular}{lccccc}
            
            \textbf{LLM} & \textbf{Method} & \textbf{Simple (\%)} & \textbf{Moderate (\%)} & \textbf{Challenging (\%)} & \textbf{Total (\%)} \\ 
            \hline
            
            \multirow{4}{*}{\textbf{Gemma-3-4B}} 
            & Ideal Schema Linking & 42.49 & 21.94 & 16.67 & 33.83 \\ \cline{2-6}
            & Baseline & 30.05 & 13.76 & 7.64 & 23.01 \\ 
            & Retrieval & 33.51 & 17.20 & 13.19 & 26.66 \\ 
            & SchemaGraphSQL$_{n-n}$ & 35.46 & 17.63 & 12.50 & 27.90 \\
            & SchemaGraphSQL$_{1-1}$ & 28.76 & 11.61 & 8.33 & 21.64 \\
            & SchemaGraphSQL$_{force-union}$ & 35.35 & 18.92 & 20.83 & \textbf{29.01} \\
            \toprule
            
            \multirow{4}{*}{\textbf{Gemma-3-12B}} 
            & Ideal Schema Linking & 58.38 & 41.08 & 29.86 & 50.46 \\ \cline{2-6}
            & Baseline & 42.59 & 22.15 & 16.67 & 33.96 \\ 
            & Retrieval & 46.38 & 30.97 & 27.08 & 39.90 \\ 
            & SchemaGraphSQL$_{n-n}$ & 52.00 & 35.05 & 27.78 & 44.59 \\
            & SchemaGraphSQL$_{1-1}$ & 50.59 & 29.03 & 23.61 & 41.53 \\
            & SchemaGraphSQL$_{force-union}$ & 54.38 & 35.27 & 26.39 & \textbf{45.96} \\ \toprule
            
            \multirow{4}{*}{\textbf{Gemma-3-27B}} 
            & Ideal Schema Linking & 63.14 & 47.96 & 38.19 & 56.19 \\ \cline{2-6}
            & Baseline & 49.41 & 31.40 & 25.69 & 41.72 \\ 
            & Retrieval & 52.22 & 41.51 & 33.33 & 47.20 \\ 
            & SchemaGraphSQL$_{n-n}$ & 59.68 & 45.16 & 34.03 & 52.87 \\ 
            & SchemaGraphSQL$_{1-1}$ & 58.38 & 41.08 & 31.94 & 50.65 \\
            & SchemaGraphSQL$_{force-union}$ & 61.19 & 44.73 & 37.50 & \textbf{53.98} \\ \toprule

            \multirow{6}{*}{\textbf{Gemini-2.5-Flash}} 
            & Ideal Schema Linking & 71.46 & 55.48 & 47.92 & 64.41 \\ \cline{2-6}
            & Baseline & 59.35 & 41.08 & 34.72 & 51.50 \\ 
            & Retrieval & 64.11 & 50.97 & 45.83 & 58.41 \\ 
            & SchemaGraphSQL$_{n-n}$ & 68.22 & 53.33 & 44.44 & 61.47 \\  
            & SchemaGraphSQL$_{1-1}$ & 66.81 & 51.61 & 43.06 & 59.97 \\  
            & SchemaGraphSQL$_{force-union}$ & 68.32 & 56.13 & 50.00 & \textbf{62.91} \\  
            \toprule

        \end{tabular}
    }
\end{table*}
\subsection{End-to-End Execution Accuracy}
\label{ssec:exec}
Table~\ref{tab:execution_accuracy_dev} reports execution accuracy for four LLM generators.  
Across the board, SchemaGraphSQL yields gains of \textbf{6–12 \%} over the single-step baseline.  
Using Gemini-2.5-Flash, \textbf{SchemaGraphSQL$_{force-union}$} attains \textbf{62.91 \%} total accuracy—only 1.5 \% short of the oracle “ideal schema linking’’ setting, implying that most residual errors stem from SQL generation rather than linking.

Improvements concentrate on the Moderate and Challenging subsets: Gemini-2.5-Flash sees a +15 \% boost on challenging questions, reflecting SchemaGraphSQL’s advantage on multi-join queries.

For every generator, the high-recall \textit{force-union} variant outperforms the high-precision 1-1 variant on execution accuracy by 2–7 \% (Dev) and 4–12 \% (MiniDev).  
This affirms that \emph{omitting} a table is far more damaging than including extras—LLMs can ignore noise but cannot guess missing joins.  
Among schema metrics, F\textsubscript{6} correlates best with end-to-end success: the highest-F\textsubscript{6} model is invariably the highest-accuracy model, whereas precision alone can be misleading.

\subsection{Efficiency}
\label{ssec:efficiency}

Our pipeline adds negligible latency: one Gemini-Flash call consumes on average 4.6 K input and 14 output tokens, and the subsequent \(O(|E|)\) shortest-path search completes in under 15 ms on commodity hardware.  
Thus SchemaGraphSQL is compatible with real-time database interfaces and low-resource deployments.

\section{Conclusion}
We have presented \textsc{SchemaGraphSQL}, a lightweight, zero-shot schema linking framework that integrates classical path-finding algorithms into modern LLM-based Text-to-SQL systems. Unlike prior work that often relies on heavy prompting techniques or supervised fine-tuning, our method outperforms prior work in schema linking with minimal computational overhead. Beyond accuracy gains, \textsc{SchemaGraphSQL} offers a transparent and interpretable mechanism for schema filtering, making it well-suited for practical deployment in real-world text-to-SQL systems.

\section*{Limitation}
While \textsc{SchemaGraphSQL} delivers strong performance on large-scale databases with well-structured foreign key relations, it has several limitations. First, our approach is not optimized for deeply nested or compositional queries that require complex subquery reasoning. Second, on dense schema graphs with excessive or noisy foreign key links, the shortest-path enumeration may yield overly broad candidate sets, affecting precision. Lastly, we treat all join paths equally and do not incorporate heuristics or weights for foreign key importance or estimated join costs, which could further improve path selection and SQL execution quality.

\bibliography{references}

\clearpage
\appendix
\section{Prompts}
\label{sec:appendix}

\label{sec:appendix-prompts}

This section includes all system prompts used throughout the SchemaGraphSQL pipeline. These prompts are designed to be modular and reusable across different configurations and model sizes.

\begin{itemize}
    \item \textbf{Prompt~\ref{prompt:joinpath}}: Selection of the most appropriate join path among candidate schema paths.
    \item \textbf{Prompt~\ref{prompt:sqlitegen}}: SQL query generation using the filtered schema and join path.
    \item \textbf{Prompt~\ref{prompt:sqlitebaseline}}: Baseline SQL generation prompt using the full schema without schema linking.
\end{itemize}

These prompts are issued via Gemini 2.5 Flash with low temperature settings to ensure stability and determinism during inference.

\refstepcounter{prompt}
\begin{tcolorbox}[
    enhanced,
    breakable,
    width=\linewidth,
    colback=gray!5,
    colframe=blue!60!black,
    fonttitle=\bfseries,
    title={\textbf{Prompt~\arabic{prompt}: System prompt for join path selection}},
    boxrule=0.6pt,
    arc=2pt,
    left=2pt, right=2pt, top=2pt, bottom=2pt
  ]
\footnotesize
\textbf{ROLE \& OBJECTIVE}\\
You are a database expert tasked with selecting the optimal \emph{join path} to answer user questions using a provided SQL schema.

\medskip
\textbf{TASK}\\
Choose the \emph{single most appropriate} join path from a list of candidates that correctly connects the relevant tables.

\medskip
\textbf{INSTRUCTIONS}
\begin{enumerate}[leftmargin=1.2em,itemsep=3pt]
  \item Internally inspect each path to determine:
    \begin{itemize}[leftmargin=*,nosep]
      \item whether it connects all necessary tables,
      \item whether joins are complete and valid,
      \item and whether it satisfies the intent of the question.
    \end{itemize}
    Briefly justify your decision \emph{internally} but \emph{do not} include any reasoning in the final output.
  \item Output one line in the following format:  
    \texttt{Final Answer: path\_id: <ID>}
\end{enumerate}
\end{tcolorbox}
\label{prompt:joinpath}

\refstepcounter{prompt}
\begin{tcolorbox}[
    enhanced,
    breakable,
    width=\linewidth,
    colback=gray!5,
    colframe=blue!60!black,
    fonttitle=\bfseries,
    title={\textbf{Prompt~\arabic{prompt}: System prompt for SQLite query generation after schema linking}},
    boxrule=0.6pt,
    arc=2pt,
    left=2pt, right=2pt, top=2pt, bottom=2pt
  ]
\footnotesize
\textbf{ROLE \& OBJECTIVE}\\
You are an expert in SQLite query generation. Your task is to generate a valid query to answer a user question based on the given schema and join path.

\medskip
\textbf{INPUTS}
\begin{itemize}[leftmargin=1.1em,itemsep=2pt]
  \item \textbf{Schema:} \verb|{schema}|
  \item \textbf{Join Path:} \verb|{join_path_string}|
  \item \textbf{Question Context:} \verb|{evidence_string}|
\end{itemize}

\medskip
\textbf{INSTRUCTIONS}
\begin{enumerate}[leftmargin=1.2em,itemsep=3pt]
  \item Use the provided schema and join path to construct a valid \texttt{SQLite} query.
  \item Ensure the query correctly answers the user's question.
  \item Format the query clearly and confirm it adheres to SQLite syntax.
\end{enumerate}
\end{tcolorbox}
\label{prompt:sqlitegen}

\refstepcounter{prompt}
\begin{tcolorbox}[
    enhanced,
    breakable,
    width=\linewidth,
    colback=gray!5,
    colframe=blue!60!black,
    fonttitle=\bfseries,
    title={\textbf{Prompt~\arabic{prompt}: Baseline prompt for SQLite query generation}},
    boxrule=0.6pt,
    arc=2pt,
    left=2pt, right=2pt, top=2pt, bottom=2pt
  ]
\footnotesize
\textbf{ROLE \& OBJECTIVE}\\
You are an expert in SQLite query generation. Your task is to produce a valid query that answers a user's question using the provided schema.

\medskip
\textbf{INPUTS}
\begin{itemize}[leftmargin=1.1em,itemsep=2pt]
  \item \textbf{Schema:} \verb|{schema}|
  \item \textbf{Question Context:} \verb|{evidence_string}|
\end{itemize}

\medskip
\textbf{INSTRUCTIONS}
\begin{enumerate}[leftmargin=1.2em,itemsep=3pt]
  \item Generate a correct \texttt{SQLite} query that answers the user question.
  \item Ensure the query is syntactically valid and aligns with the schema.
  \item Format the query clearly and cleanly.
\end{enumerate}
\end{tcolorbox}
\label{prompt:sqlitebaseline}

\section{Additional Results}
\label{sec:appendix-results}

This section presents extended evaluation results that complement those in the main text. We report schema linking scores and execution accuracy on the \textsc{MiniDev} split of the BIRD dataset to validate robustness and generalization.

\begin{itemize}
    \item \textbf{Table~\ref{tab:schema_linking_minidev}}: Comparison of schema linking methods on MiniDev, including LLM baselines, dense retrievers, and SchemaGraphSQL.
    \item \textbf{Table~\ref{tab:schema_linking_results_methods_minidev}}: SchemaGraphSQL ablation results across different graph configurations on MiniDev.
    \item \textbf{Table~\ref{tab:execution_accuracy_minidev}}: End-to-end SQL execution accuracy for all models and schema linking variants on MiniDev, broken down by question difficulty.
\end{itemize}

These extended results reinforce the strong recall and execution performance of SchemaGraphSQL, especially on complex and multi-table SQL queries.

\begin{table*}[h]
    \centering
    \caption{Schema Linking Results in MiniDev Dataset}
    \label{tab:schema_linking_minidev}
    \resizebox{\linewidth}{!}{
        \begin{tabular}{|l|ccccc|}
            \hline
            \textbf{Method} & \textbf{Exact Match Rate (\%)} & \textbf{Precision (\%)} & \textbf{Recall (\%)} & \textbf{F1 (\%)} & \textbf{F6 (\%)} \\ \hline
            LLM as Schema Linker & 75.70 & 92.82 & 90.56 & 91.68 & 90.62 \\ 
            Retrieval (Top1) & 14.40 & 86.40 & 41.24 & 55.83 & 41.83 \\ 
            Retrieval (Top2) & 28.00 & 68.30 & 64.67 & 66.43 & 64.76 \\ 
            Retrieval (Top3) & 4.80 & 55.00 & 77.73 & 64.42 & 76.88 \\ 
            Retrieval (Top4) & 1.00 & 47.29 & 85.00 & 60.77 & 83.20 \\ 
            Retrieval (Top5) & 0.80 & 41.52 & 89.64 & 56.75 & 86.92 \\ 
            Retrieval (Top6) & 0.80 & 37.06 & 92.26 & 52.87 & 88.69 \\ 
            SchemaGraphSQL (Ours) & 82.33 & 94.80 & \textbf{93.97} & \textbf{94.38} & \textbf{93.99} \\ 

\hline
        \end{tabular}
    }
\end{table*}

\begin{table*}[h!]
	\centering
	\caption{Schema Linking Results Across Different Graph Settings (Minidev)}
	\label{tab:schema_linking_results_methods_minidev}
	\resizebox{\linewidth}{!}{
		\begin{tabular}{|l|ccccc|}
			\hline
			\textbf{Method} & \textbf{Exact Match Rate (\%)} & \textbf{Precision (\%)} & \textbf{Recall (\%)} & \textbf{F1 (\%)} & \textbf{F6 (\%)} \\ \hline
SchemaGraphSQL$_{1-1}$ & 64.86 & \textbf{96.47} & 79.67 & 87.27 & 80.05 \\ 
SchemaGraphSQL$_{1-n}$ & 74.10 & 95.93 & 87.16 & 91.34 & 87.38 \\ 
SchemaGraphSQL$_{n-1}$ & 82.13 & 95.81 & 93.39 & \textbf{94.58} & 93.45 \\ 
SchemaGraphSQL$_{n-n}$ & \textbf{82.33} & 94.80 & 93.97 & 94.38 & 93.99 \\ 
SchemaGraphSQL$_{force-longest}$ & 72.29 & 92.97 & 86.19 & 89.45 & 86.36 \\ 
SchemaGraphSQL$_{no-union}$ & 74.90 & 95.16 & 87.94 & 91.41 & 88.12 \\ 
SchemaGraphSQL$_{force-union}$ & 80.72 & 89.36 & \textbf{94.75} & 91.97 & \textbf{94.59} \\ 

\hline
		\end{tabular}
	}
\end{table*}

\begin{table*}[h!]
    \centering
    \caption{SQL Execution Accuracy Results - MiniDev}
    \label{tab:execution_accuracy_minidev}
    \resizebox{\linewidth}{!}{
        \begin{tabular}{lccccc}
            
            \textbf{LLM} & \textbf{Method} & \textbf{Simple (\%)} & \textbf{Moderate (\%)} & \textbf{Challenging (\%)} & \textbf{Total (\%)} \\ 
            \hline
            
            \multirow{4}{*}{\textbf{Gemma-3-4B}} 
            & Ideal Schema Linking & 47.97 & 21.37 & 18.63 & 28.71 \\ \cline{2-6}
            & Baseline & 32.43 & 10.08 & 6.86 & 16.06 \\ 
            & Retrieval & 36.49 & 18.55 & 13.73 & 22.89 \\ 
            & SchemaGraphSQL$_{n-n}$ & 42.57 & 18.15 & 12.75 & 24.3 \\
            & SchemaGraphSQL$_{1-1}$ & 31.08 & 10.08 & 6.86 & 15.66 \\
            & SchemaGraphSQL$_{force-union}$ &  41.89 & 18.95 & 15.69 & \textbf{25.1} \\ \toprule
            
            \multirow{4}{*}{\textbf{Gemma-3-12B}} 
            & Ideal Schema Linking & 63.51 & 45.56 & 34.31 & 48.59 \\ \cline{2-6}
            & Baseline & 38.51 & 18.95 & 16.67 & 24.3 \\ 
            & Retrieval & 50.68 & 35.08 & 28.43 & 38.35 \\ 
            & SchemaGraphSQL$_{n-n}$ & 57.43 & 37.5 & 30.39 & 41.97 \\
            & SchemaGraphSQL$_{1-1}$ & 54.73 & 29.03 & 23.53 & 35.54 \\
            & SchemaGraphSQL$_{force-union}$ & 60.14 & 42.34 & 33.33 & \textbf{45.78} \\ \toprule
            
            \multirow{4}{*}{\textbf{Gemma-3-27B}} 
            & Ideal Schema Linking & 72.97 & 53.63 & 43.14 & 57.23 \\ \cline{2-6}
            & Baseline & 50.0 & 27.82 & 21.57 & 33.13 \\ 
            & Retrieval & 60.81 & 44.76 & 36.27 & 47.79 \\ 
            & SchemaGraphSQL$_{n-n}$ & 66.22 & 50.81 & 35.29 & 52.21 \\
            & SchemaGraphSQL$_{1-1}$ & 61.49 & 38.71 & 27.45 & 43.17 \\
            & SchemaGraphSQL$_{force-union}$ & 68.92 & 52.02 & 44.12 & \textbf{55.42} \\ \toprule

            \multirow{6}{*}{\textbf{Gemini-2.5-Flash}} 
            & Ideal Schema Linking & 83.78 & 66.13 & 56.86 & 69.48 \\
 \cline{2-6}
            & Baseline & 58.78 & 43.95 & 36.27 & 46.79 \\ 
            & Retrieval & 75.0 & 53.63 & 53.92 & 60.04 \\ 
            & SchemaGraphSQL$_{n-n}$ & 77.03 & 58.87 & 50.98 & 62.65 \\
            & SchemaGraphSQL$_{1-1}$ & 76.35 & 56.85 & 41.18 & 59.44 \\
            & SchemaGraphSQL$_{force-union}$ & 77.7 & 62.5 & 50.98 & \textbf{64.66} \\ \toprule
            
        \end{tabular}
    }
\end{table*}

\end{document}